# Enhancing hierarchical surrogate-assisted evolutionary algorithm for high-dimensional expensive optimization via random projection

Xiaodong Ren [1], Daofu Guo [1], Zhigang Ren [1], Yongsheng Liang [1], An Chen [1]

[1]School of Automation Science and Engineering, Xi'an Jiaotong University, Xi'an 710049, China

**Corresponding author**: Zhigang Ren, School of Automation Science and Engineering, Xi'an Jiaotong University, Xi'an 710049, China

**E-mail**: renzg@xjtu.edu.cn

**Acknowledgements**

This work was supported in part by the National Natural Science Foundation of China under Grants 61873199 and in part by the Natural Science Basic Research Plan in Shaanxi Province of China under Grants 2020JM-059. Besides, the authors would like to thank Xinjing Wang, Chaoli Sun, and Haibo Yu for providing source code of ESAO, SA-COSO, and SHPSO, respectively.

**Abstract**: By remarkably reducing real fitness evaluations, surrogate-assisted evolutionary algorithms (SAEAs), especially hierarchical SAEAs, have been shown to be effective in solving computationally expensive optimization problems. The success of hierarchical SAEAs mainly profits from the potential benefit of their global surrogate models known as "blessing of uncertainty" and the high accuracy of local models. However, their performance leaves room for improvement on high-dimensional problems since now it is still challenging to build accurate enough local models due to the huge solution space. Directing against this issue, this study proposes a new hierarchical SAEA by training local surrogate models with the help of the random projection technique. Instead of executing training in the original high-dimensional solution space, the new algorithm first randomly projects training samples onto a set of low-dimensional subspaces, then trains a surrogate model in each subspace, and finally achieves evaluations of candidate solutions by averaging the resulting models. Experimental results on six benchmark functions of 100 and 200 dimensions demonstrate that random projection can significantly improve the accuracy of local surrogate models and the new proposed hierarchical SAEA possesses an obvious edge over state-of-the-art SAEAs.

# 1. Introduction

Evolutionary algorithms (EAs), such as differential evolution (DE) [1], genetic algorithm (GA) [2], and particle swarm optimization (PSO) [3], have been widely employed to solve real-world engineering optimization problems [4-6]. These EAs generally require a large number of fitness evaluations (FEs) to find a satisfying solution, which make them unsuitable for computationally expensive optimization problems [7-10]. The reason consists in that a single FE of these expensive problems often consumes lots of time or material resources.

Surrogate model provides an effective tool to reduce computational cost by partly replacing computationally expensive FEs during the evolution process. Over the past decades, several types of surrogate models, including polynomial regression [11], radial basis function (RBF) [12-14], Gaussian process (GP) [15, 16], and support vector machine [17], have been developed and deeply analyzed, yielding various surrogate-assisted evolutionary algorithms (SAEAs). According to the role of the employed surrogate model, existing SAEAs can be roughly divided into the following three categories: 1) global SAEAs, 2) local SAEAs, and 3) hierarchical SAEAs. Global surrogate models aim to approximate the expensive fitness function in the whole solution space [18-22]. As a contrast, local surrogate models focus on a small solution region for the purpose of ensuring the approximation accuracy [23-26]. Fusing the exploration capability of the global surrogate and the exploitation capability of the local surrogate together, hierarchical SAEAs attracted much research attention in recent years and showed certain superiority over a single type of SAEAs on most expensive problems [27-30].

So far, SAEAs have achieved great success in tackling low- and medium-dimensional expensive problems, but they generally lose efficiency on high-dimensional problems due to the "curse of dimensionality" [31]. On the one hand, EAs cannot fully explore the huge solution space of a high-dimensional problem at the cost of acceptable computational resources. On the other hand, it is also impracticable to directly build an accurate enough surrogate model with a limited number of available training samples. Directing at the above two challenges, several beneficial attempts have been made. Tian *et al.* [32] developed a multiobjective sample infill criterion according to non-dominated sorting, and enhanced the adaptability of a GP-assisted PSO algorithm. Li *et al.* [33] proposed a surrogate-assisted multiswarm optimization algorithm, where a swarm is specially evolved to enhance the exploration capability of the whole algorithm.

Besides, it has been verified that hierarchical SAEAs are of great potential in solving high-dimensional expensive problems [34-37]. Profiting from its "blessing of uncertainty" [38], the global surrogate model generally helps to smooth out some local optima and thus to reduce the search space, whereas the local surrogate model helps to identify better solutions in the located local promising regions. Taking a recently proposed hierarchical SAEA called evolutionary sampling assisted optimization (ESAO) [34] as an example, it builds a global RBF model to assist DE to conduct global search by prescreening promising solutions. On the other side, it performs local search by taking another RBF model trained in the neighborhood of the current best solution as the objective function. ESAO alternates the two types of search if one of them cannot lead to a better solution.

By this means, it not only enhances the possibility of finding the global optimum but also speeds up the optimization process. Compared with the global surrogate model, the local model in a hierarchical SAEA is expected to be of much higher accuracy. Despite the reduced solution region, it is still a nontrivial task to build an accurate enough local surrogate model in a high-dimensional solution space.

To alleviate this issue, this study proposes a random projection-enhanced hierarchical SAEA (RPHSA). RPHSA inherits the fine framework of ESAO, but adapts the local surrogate model therein, i.e., RBF, to high-dimensional problems with the random projection (RP) technique. As a commonly-used dimension reduction technique, RP can expediently project high-dimensional data onto subspaces of much lower dimensions while maintaining the geometric structure among the data to a great extent. Now, RP has been successfully applied in many fields such as signal processing [39], machine learning [40], and high-dimensional optimization [41], but has seldom been employed to train surrogate models for high-dimensional problems. With the introduction of RP, RPHSA builds its local RBF model as follows: first randomly project original training samples onto several low-dimensional subspaces, then train low-dimensional RBF models in respective subspaces so as to capture the characteristics of the original problem from different perspectives, and finally construct the final RP-based RBF (RP-RBF) by averaging all low-dimensional RBF models. In this way, the final RP-RBF is expected to achieve ideal approximation capability with a small number of training samples and thus to enhance the performance of RPHSA.

The remainder of this paper is organized as follows. Section 2 briefly reviews the related work of SAEAs. Section 3 describes the proposed RP-RBF model and the resulting RPHSA algorithm in detail. Section 4 reports experimental settings and results along with some analyses. The conclusion is finally given in Section 5.

## 2. Related Work

SAEAs have been receiving more and more research attention in recent years for their effectiveness in solving computationally expensive problems. They replace most of the expensive real FEs with surrogate estimations in the optimization process, such that many computational resources can be avoided and the optimization performance can be greatly improved. Early SAEAs tend to employ global surrogate models to fit the whole optimization problem. Their excellent performance on simple and low-dimensional expensive optimization problems have been verified. Jin *et al.* [18] adopted an artificial neural network as a global surrogate to assist the covariance matrix adaptation evolution strategy and also proposed an empirical criterion to switch between expensive real FEs and cheap fitness estimations during the evolutionary search process. Ratle [19] suggested using GP to construct a global surrogate so as to guide the search process. Regis *et al.* [20] developed a SAEA based on PSO and a global RBF model, where the former generates multiple trial positions for each particle in each iteration, while the later prescreens most promising trial positions to generate new particles. Liu *et al.* [21] employed a GP with lower confidence bound to select promising solutions in the evolution process of a DE algorithm. They also utilized

Sammon mapping, a dimension reduction technique, to enhance the surrogate accuracy on medium-scale expensive optimization. Dong et al. [22] proposed a multi-start approach with the goal of finding all the local optima of a pre-trained global GP model. And the search was performed within the located local optimal solution regions.

Global surrogate models take effect on expensive problems of simple fitness landscape, but they suffer from "curse of dimensionality" and cannot adapt themselves well to complicated problems. To alleviate this issue, researchers developed local surrogate models to improve the approximation accuracy in local solution regions. Ong et al. [23] employed a trust-region method for interleaving use of exact models for the objective and constraint functions with computationally cheap RBF models in the local search process. The idea of fitness inheritance suggested by Smith et al. [24] can also be seen as a local surrogate, where the fitness value of an individual is estimated based on its neighbors and parents. Similarly, Sun et al. [25] proposed a fitness estimation strategy to approximate particles in PSO based on their positional relationship with other particles. Lu and Tang [26] integrated a local surrogate model into DE, where the model was used not only for regression but also for classification.

Compared with global surrogate models, local models are more likely to improve the solution quality, but they tend to lack the capability of jumping out of local optima. To achieve complementary advantages of these two types of models, many studies in recent years focused on developing hierarchical SAEAs by integrating global and local surrogate together. Zhou et al. [27] proposed a hierarchical SAEA, where a global GP and a local RBF network were jointly employed to assist a GA. The former was used to identify promising individuals at the global search level, while the later was adopted to accelerate the convergence of a trust-region enabled search strategy at the local search level. Tenne and Armfield [28] proposed a memetic optimization framework consisting of variable global and local surrogates, and employed RBF in a trust-region approach for expensive optimization. Sun et al. [29] introduced a two-layer surrogate-assisted PSO, where a global surrogate model was intended to smooth out some local optima of the original multimodal fitness functions and a local surrogate model was employed for fitness estimation. Inspired by committee-based active learning, Wang et al. [30] proposed an ensemble of several global surrogates to search for the best and most uncertain solutions to be evaluated by the real fitness function, and employed a local surrogate to model the neighborhood of the current best solution with the goal of further improving it by optimizing the model.

As for high-dimensional expensive problems, they have become a research hotspot in the field of SAEAs. In addition to the ESAO algorithm introduced in Section 1, some other significant research efforts have been made to push the boundary of SAEAs in solving this kind of problems. Sun et al. [35] proposed a surrogate-assisted cooperative swarm optimization algorithm, where an RBF-assisted social learning PSO focuses on exploration and a fitness estimation strategy-assisted PSO concentrates on local search. Yu et al. [36] embedded the social learning PSO into an RBF-assisted PSO framework. The former aims to find the optimum of the RBF model, which is constructed with a certain number of the best solutions found so far, and thereby refines its local approximation of the fitness landscape around the optimum. The latter conducts search in a wider solution region, enabling the RBF model to capture the global landscape of the fitness function. Yang *et al.* [37] developed a two-layer surrogate-assisted DE algorithm. This algorithm measures its evolutionary status according to the improvement times of the best solution. According to the feedback status, three different DE mutation operators are employed to generate new offspring, which are further prescreened by a global or local GP model. Tian *et al.* [32] revealed that the approximation uncertainty of GP becomes less reliable on high-dimensional problems and the commonly-used scalar sample infill criteria, which combines the approximated fitness and the approximation uncertainty in a scalar function, tends to lose efficacy. To overcome this defect, they developed a multiobjective sample infill criterion by considering the above two factors as separate objectives and selecting promising solutions according to non-dominated sorting, and thereby achieved a good balance between exploitation and exploration. Li *et al.* [33] proposed a SAEA involving two swarms, where the first one uses the learner phase of teaching-learning-based optimization to enhance exploration and the second one uses PSO for faster convergence. Moreover, they also designed a novel sample infill criterion by selecting particles predicted to be with self-improvement for real FEs.

These recent studies enhance the adaptability of SAEAs on high-dimensional expensive problems. However, it is still an open problem to build accurate enough surrogate models, especially local models in hierarchical SAEAs, for high-dimensional problems. This study attempts to tackle this issue with the RP technique.

## 3. Proposed RPHSA

This section first introduces how to scale up RBF to high-dimensional problems with the RP technique, then discusses the integration of the resulting RP-RBF model and the basic ESAO algorithm, and finally presents the implementation of the proposed RPHSA algorithm.

### 3.1 RP-RBF model

As a commonly-used surrogate model, RBF has been shown to fit nonlinear functions well and be capable of both local and global modeling [20, 23, 34-36]. The RBF model used in this paper has an interpolation form and can be formulated as follows.

Given a set $X = (x_1, x_2, \cdots, x_n) \in \mathbb{R}^{d \times n}$ of $n$ distinct training samples in $\mathbb{R}^d$ with their fitness values being $f(x_1), f(x_2), \cdots, f(x_n)$, RBF approximates the fitness value of a new solution as

$$\hat{f}(x) = \sum_{i=1}^{n} \omega_i \varphi(\|x - x_i\|), \tag{1}$$

where $\|\cdot\|$, $\varphi(\cdot)$, and $\omega_i$ denote the Euclidean norm, the basis function, and the weight coefficient to be learnt, respectively. There are several types of basis functions, including multiquadric function, Gaussian function, and splines. In this study, the multiquadric function $\phi(r)=\sqrt{r^2+c^2}$ with the bias term $c$ being 1 is used to construct RBF. Compared with other types of surrogate models such as GP, RBF is relatively easy to train. Let $\Phi$ be an $n\times n$ matrix with each element $\Phi_{ij}=\varphi(\|x_i-x_j\|)$ and $f$ denote the vector $(f(x_1),f(x_2),\cdots,f(x_n))^T$, then the unknown coefficients $\omega=(\omega_1,\omega_2,\cdots,\omega_n)^T$ of RBF can be obtained by solving the following linear equation:

$$\omega=\Phi^{-1}f. \qquad (2)$$

Although the performance of RBF is relatively insensitive to problem dimension, its accuracy on high-dimensional problems is still difficult to be guaranteed due to the huge modeling space and the very limited number of available training samples. To cope with this dilemma, a promising way is to reduce the modeling space with a dimension reduction technique. This study selects RP to execute this task for its simplicity and capability of preserving the geometric structure among training samples [42].

To perform projection with RP, a random projection matrix independent of training data is required. According to [43], the Gaussian random projection matrix, each of whose elements obeys the standard normal distribution, meets this requirement. To ensure the orthogonality required by RP, we further perform column orthogonalization on the initially generated Gaussian matrix. Let $P\in\mathbb{R}^{k\times d}$ denote the final projection matrix with $d$ and $k$ being space dimensions before and after projection, respectively, then a set of low-dimensional training samples can be obtained through the following projection operation:

$$X'=PX=(x'_1,x'_2,\cdots,x'_n)\in\mathbb{R}^{k\times n}. \qquad (3)$$

Taking $X'=(x'_1,x'_2,\cdots,x'_n)$ as training samples, a low-dimensional RBF model $\hat{f}(Px)$ can be generated according to (1) and (2).

The excellent dimension reduction property of RP can be theoretically guaranteed by the following Johnson-Lindenstrauss Lemma [43]:

**Lemma 1.** (Johnson-Lindenstrauss Lemma): Given $0<\varepsilon<1$, a set $X$ of $n$ points in $\mathbb{R}^d$, and a positive integer $k$ satisfying $k\geq k_0=O(\log(n)/\varepsilon^2)$, there exists a linear map $P:\mathbb{R}^d\to\mathbb{R}^k$ such that

$$(1-\varepsilon)\|x_i-x_j\|^2\leq\|P(x_i)-P(x_j)\|^2\leq(1+\varepsilon)\|x_i-x_j\|^2 \qquad (4)$$

for $\forall x_i, x_j\in X$.

This lemma suggests that the distance between any two points in high-dimensional Euclidean space, i.e., the geometric structure of the original training data, can be preserved after projection within a certain error range, which depends on the dimension of the new space. The lower the new dimension is, the more greatly the difficulty of training an RBF can be reduced, but the broader the error range tend to be, which is harmful to the accuracy of the trained RBF model. To balance this contradiction, the developed RP-RBF model first projects the original training samples onto a group of low-dimensional random subspaces instead of a single one, and then trains an RBF in each subspace. In this way, the original training samples can be shared in different subspaces and a low-dimensional RBF capturing part of characteristics of the original problem can be easily trained in each subspace. By averaging all low-dimensional RBFs, the final RP-RBF is expected to be able to learn more characteristics of the original problem and to achieve higher accuracy. Fig. 1 illustrates the construction process of RP-RBF. With $m$ being the number of subspaces and $P_i \in \mathbb{R}^{k \times d}$ being the $i$th random projection matrix, the final RP-RBF model can be formulated as

$$\hat{f}(\boldsymbol{x}) \triangleq \sum_{i=1}^{m} \hat{f}_i(P_i \boldsymbol{x}) \Big/ m. \tag{5}$$

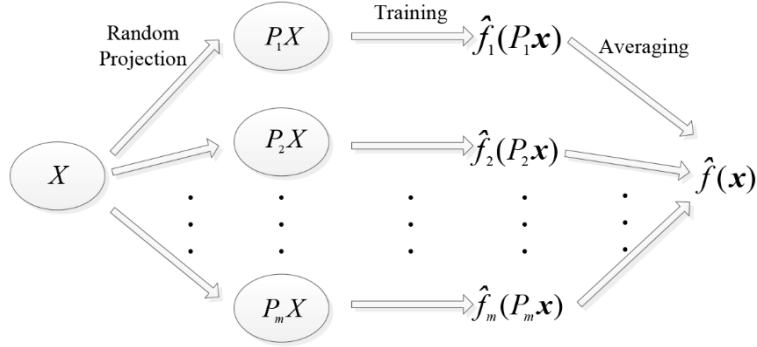

**Fig. 1** Construction process of RP-RBF

Algorithm 1 presents the pseudocode for constructing RP-RBF, where steps 2-4 show the specific process of building an RBF in each subspace. It is notable that the subspace dimension $k$, the subspace number $m$, and the sample number $n$ have an important influence on the accuracy of RP-RBF. We will empirically analyze their influence in the experimental section.

---
**Algorithm 1**: Pseudocode for constructing RP-RBF

**Input:** training samples $X \in \mathbb{R}^{d \times n}$ with corresponding fitness vector $\boldsymbol{f} \in \mathbb{R}^n$, subspace dimension $k$, and subspace number $m$;
**Output:** RP-RBF model $\hat{f}(\boldsymbol{x})$.

1. **for** $i = 1$ to $m$
2.     Generate a random projection matrix $P_i \in \mathbb{R}^{k \times d}$ by performing column orthogonalization on a Gaussian matrix;
3.     Generate the $i$th group of training samples $X_i'$ by performing projection with $P_i$ according to (3);
4.     Construct an RBF model $\hat{f}_i(P_i \boldsymbol{x})$ based on $X_i'$ and $\boldsymbol{f}$ according to (1) and (2);
5. Generate the final RB-RBF model $\hat{f}(\boldsymbol{x})$ by averaging all low-dimensional models according to (5).

---

## 3.2 Integration of RP-RBF and ESAO

As a recently developed hierarchical SAEA, ESAO maintains a global and a local RBF model to assist the global and the local search conducted by DE, respectively [34]. The global RBF is trained in each generation of DE with all truly evaluated solutions. It is employed to predict the fitness values of all offspring generated by DE through mutation and crossover operator. The offspring with the lowest prediction will be further evaluated with the real fitness function, and will replace its parent if it is better. ESAO will also employ this offspring to update the current best solution and continue the global search process if it is of better fitness, otherwise ESAO will switch to the local search process. The local RBF is trained with a certain number of best solutions that have been truly evaluated. ESAO takes it as an approximate fitness function and employs DE to find its optimum. This optimum will undergo real evaluation. If it is better than the current best solution, ESAO will start a new local process after updating the current best solution with it and adding it to the population of the global process, otherwise ESAO will return to the global search process.

A straightforward way to integrate RP-RBF and ESAO is to build both the global and local surrogate model for ESAO with RP-RBF. However, the proposed RPHSA tends to just replace the local model with RP-RBF. The reasons are threefold. First, the local RBF model aims to capture fitness landscape details of the current promising local solution region. It is directly used as the fitness function for local search and thereby requires much on estimation accuracy. On the contrary, the global RBF model is designed to describe all explored solution regions. It is allowed to be of certain estimation error such that some local optima can be smoothed out. Second, the structural characteristics of a local high-dimensional solution region are relatively simple and are more likely to be preserved and learned in low-dimensional subspaces. As a contrast, the structural characteristics of the solution regions covered by the global RBF model are much more complicated and can hardly be modeled with high enough accuracy. Finally, experimental results in [34] reveal that more real FEs are conducted in the local search process than in the global one, which means that ESAO depends more on the local search process in seeking high-quality solutions. Therefore, it is hopeful to achieve better optimization performance by adopting RP-RBF as the local surrogate model in ESAO.

Taking the basic framework of ESAO, Algorithm 2 presents the pseudocode of RPHSA. As suggested by ESAO, step 1 initializes the population for global search by optimal Latin hypercube sampling (OLHS) [44]. Steps 3-13 execute the global search, and steps 14-22 conducts the local search. The two types of search alternate if they fail to find a better solution. RPHSA terminates its optimization process when meeting a stopping criterion, which is generally set as a maximum number of real FEs.

## 4. Experimental Studies

To investigate the effectiveness and efficiency of RPHSA, we tested it on six widely-used benchmark functions and empirically compared it with ESAO and some other state-of-the-art SAEAs for high-dimensional expensive problems. Table 1 lists the characteristics of these benchmark functions. They can all be set to 100 dimensions (100-D) or 200 dimensions (200-D). In our experiments, each algorithm was required to independently run 30 times on each function, and the maximum number of real FEs for each run was set to 1000 for both 100-D and 200-D functions. The results reported below are all average ones on 30 runs.

**Algorithm 2**: Pseudocode of RPHSA

**Input**: subspace dimension $k$ and subspace number $m$;

**Output**: the current best solution $x_b$ and its true fitness function value $f(x_b)$.

1. Generate the initial population using OLHS and evaluate each individual therein with the real function $f(x)$;
2. Initialize the current best solution $x_b$ with the best solution in the current population;
3. **While** stopping criterion is not met **do**
4.     Generate offspring by performing mutation and crossover operations of DE on the current population;
5.     Build a global RBF model $\hat{f}_{global}(x)$ with all truly evaluated solutions;
6.     Predict the fitness values of all offspring using $\hat{f}_{global}(x)$;
7.     Find out the offspring with the lowest prediction $x_g$ and evaluate it with $f(x)$;
8.     **if** $f(x_g) < f(x_{gp})$ with $x_{gp}$ being the parent of $x_g$
9.         Replace $x_{gp}$ with $x_g$;
10.     **if** $f(x_g) < f(x_b)$
11.         Replace $x_b$ with $x_g$;
12.     **else**
13.         Go to step 14;
14. **While** stopping criterion is not met **do**
15.     Find out $n$ best solutions that have been truly evaluated;
16.     Build a local RBF model $\hat{f}_{local}(x)$ by running **Algorithm 1**;
17.     Locate the optimum $x_l$ of $\hat{f}_{local}(x)$ by taking DE as optimizer;
18.     Evaluate $x_l$ with $f(x)$;
19.     **if** $f(x_l) < f(x_b)$
20.         Replace $x_b$ with $x_l$ and add $x_l$ to the population for global search;
21.     **else**
22.         Go to step 3.

**Table 1** Characteristics of benchmark functions

| Function | Description | Property | Global optimum |
|---|---|---|---|
| F1 | Ellipsoid | Unimodal | 0 |
| F2 | Rosenbrock | Multimodal with narrow valley | 0 |
| F3 | Ackley | Multimodal | 0 |
| F4 | Griewank | Multimodal | 0 |
| F5 | Shifted Rotated Rastrigin | Very complicated multimodal | -330 |
| F6 | Rotated hybrid Composition Function | Very complicated multimodal | 10 |

## 4.1 Influence of parameters

Compared with ESAO, RPHSA introduces two new parameters, i.e., the dimension ($k$) and number ($m$) of subspaces used for training RP-RBF. As revealed by Lemma 1 in Section 3.1, the value of $k$ depends on the number ($n$) of training samples to be projected and the allowed projection error. A large value of $k$ is beneficial to keeping the geometric structure among training samples and thus to reducing projection error, but simultaneously limits the difficulty reduction of training an RBF. Therefore, we tested the joint influence of $k$ and $n$ on the performance of RPHSA. The candidate values $k \in \{20, 30, 40, 50, 60\}$ and $n \in \{50, 100, 150, 200\}$ were considered in the experiment. As for $m$, a proper value is helpful to preserve characteristics of the original problem. However, a too large value of $m$ will increase the computational burden for building RP-RBF. According to the recommendation in [41], we directly set $m=4\lceil d/k \rceil$ as it is large enough to ensure the effectiveness of RP. The other parameters of RPHSA were strictly set to the same values as the corresponding ones in ESAO.

Fig. 2 shows the performance variation of RPHSA with respect to different combinations of *k* and *n*. For brevity, four benchmark functions, including F1-F4 of 100-D, are taken as examples here. It can be seen from Fig. 2 that *k* and *n* show a similar joint influence on the performance of RPHSA on different functions. In general, a larger value of *k* and a smaller value of *n* help PRHSA to achieve better performance, and the combination of $k=50$ and $n=100$ reaches a good trade-off among different functions.

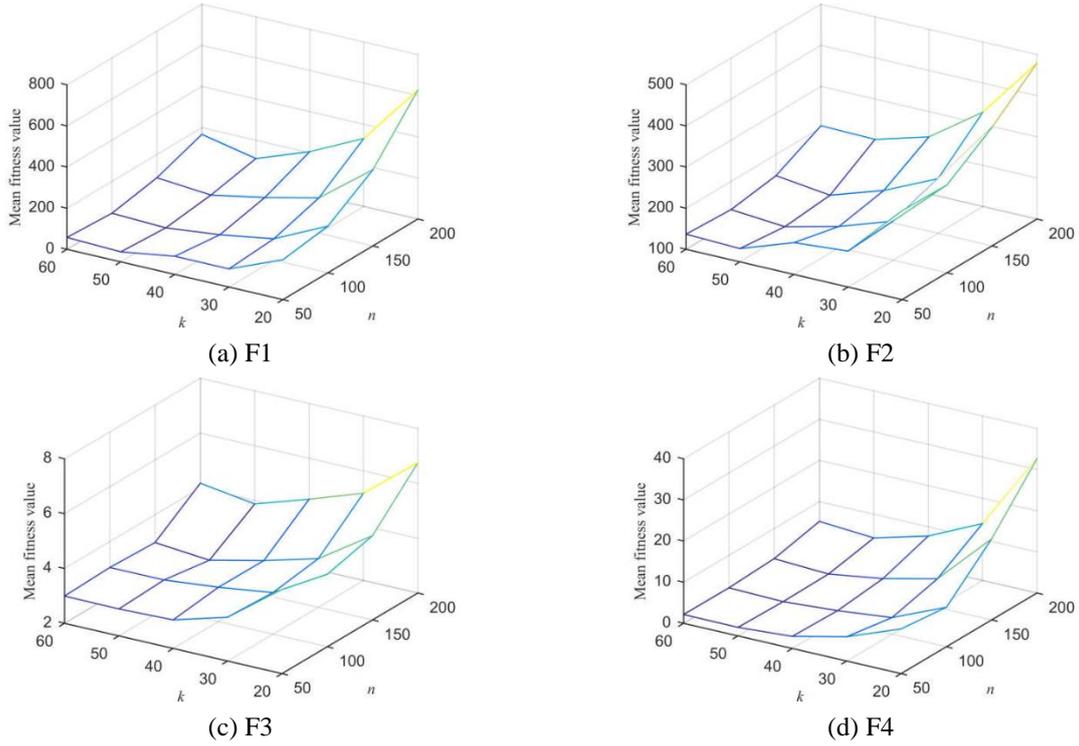

**Fig. 2** Mean fitness values obtained by RPHSA with different combinations of *k* and *n* on 100-D functions

To be specific, the RP-RBF model in RPHSA requires much fewer training samples than the traditional RBF models, which generally need at least 2*d* samples [34]. This merit mainly profits from the dimension reduction capability of RP. The unappealing performance of RPHSA with a large value of *n* may consist in the following two reasons: On the one hand, too many training samples either concern more than one local solution region or cause overfitting in a single local solution region. On the other hand, it is almost impractical for RP to simultaneously keep the geometric structure among a large number of training samples. As for *k*, when it is set to a value from {40, 50, 60}, RPHSA demonstrates similar and acceptable performance. However, the performance of RPHSA significantly deteriorates when *k* becomes smaller. This is understandable because a too small value of *k* will make the original high-dimensional training samples lose too much structural information during projection, which will further reduce the approximation accuracy of RP-RBF. We also tested RPHSA with different combinations of *k* and *n* on 200-D functions. It was found that the combination of $k=50$ and $n=100$ also enables RPHSA to get superior results. Therefore, we set this combination as the default setting of RPHSA and employ it in the following experiments.

## 4.2 Effectiveness of RP-RBF

To verify the effectiveness of the developed RP-RBF model, we empirically compared it with the traditional RBF model in terms of approximation accuracy and capability of enhancing optimization performance.

1) Approximation accuracy. In this experiment, we specially built a traditional local RBF model in RPHSA besides RP-RBF. This model was strictly built according to the method described in ESAO. It did not participate in any algorithmic operation but being used for comparison. In the middle or late evolution stage of RPHSA, we picked out a population of DE that conducts local search and evaluated each individual therein by real fitness function, the traditional RBF, and RP-RBF. Fig. 3 presents the evaluation results, where the abscissa indicates individuals sorted in ascending order according to their real fitness values.

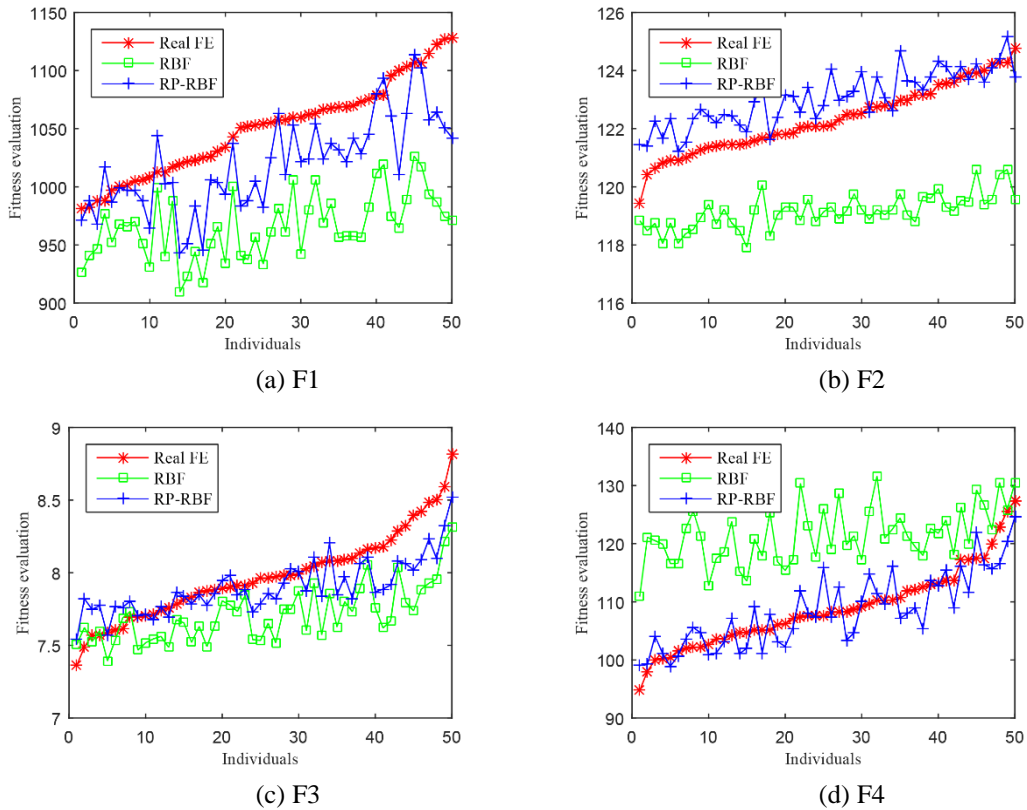

(a) F1  (b) F2

(c) F3  (d) F4

**Fig. 3** Fitness evaluation of individuals in a population of DE that conducts local search on 100-D functions

As we can see from Fig. 3, the curve of RP-RBF is more similar with the one of real FE, and it is also closer to the latter. This indicates that RP-RBF has higher approximation accuracy than the traditional RBF and thus is more likely to help the optimizer find high-quality solutions. The superiority of RP-RBF also verifies the effectiveness of RP in reducing the burden of high-dimensional modeling.

2) Optimization performance. We verified the superiority of RP-RBF in enhancing optimization performance by directly comparing the optimization results of RPHSA and ESAO since the difference of these two algorithms only lies in the local surrogate model. Table 2 reports the results on six 100-D functions, where the mean, the standard deviation, and the result of

Wilcoxon rank sum test at a significance level of 0.05 are reported. The results in bold indicate that they are the better ones, and the symbols "+", "-" and "=" indicate that the result of RPHSA is better than, worse than, and similar to the corresponding one of ESAO, respectively. It can be observed from Table 2 that RPHSA significantly outperforms ESAO on all benchmark functions except F5. It improves the results of ESAO on F1, F3, and F4 by at least an order of magnitude in terms of the mean.

Table 2 Optimization results of ESAO and RPHSA on 100-D functions

| Function | ESAO | | RPHSA | | Wilcoxon test |
|---|---|---|---|---|---|
| | Mean | Std. | Mean | Std. | |
| F1 | 1.2829E+03 | 1.3439E+02 | **3.4789E+01** | **1.1314E+01** | + |
| F2 | 5.7884E+02 | 4.4767E+01 | **1.2014E+02** | **2.1860E+01** | + |
| F3 | 1.0364E+01 | **2.1130E-01** | **3.0601E+00** | 2.9360E-01 | + |
| F4 | 5.7342E+01 | 5.8387E+00 | **1.7949E+00** | **2.5087E-01** | + |
| F5 | **7.1347E+02** | **2.6454E+01** | 7.8263E+02 | 6.7187E+01 | - |
| F6 | 1.3724E+03 | 2.7539E+01 | **1.3502E+03** | **2.0771E+01** | + |

To gain a further insight into the difference between RP-RBF and the traditional RBF, we specially calculated the total number of local searches (NLS), the number of true improvements (NTI) of local searches, and their ratios when conducting RPHSA and ESAO. Table 3 reports the average result on 30 independent runs on each 100-D function. It indicates that RPHSA has an obvious edge over ESAO in terms of all three indicators on all tested function. In particular, RPHSA outperforms ESAO by several times in terms of NTI and NTI/NLS on the first four functions, which is consistent with the results shown in Table 2. These results reveal that the local search of RPHSA is more efficient than that of ESAO, which benefits from the higher approximation accuracy of RP-RBF.

Table 3 Average NLS, NTI, and NTI/NLS of ESAO and RPHSA on 100-D functions

| Function | ESAO | | | RPHSA | | |
|---|---|---|---|---|---|---|
| | NLS | NTI | NTI/NLS | NLS | NTI | NTI/NLS |
| F1 | 414.3 | 30.1 | 7.3% | 459.4 | 120.0 | 26.1% |
| F2 | 416.3 | 33.9 | 8.1% | 461.3 | 123.7 | 26.8% |
| F3 | 414.8 | 31.2 | 7.5% | 438.1 | 77.4 | 17.7% |
| F4 | 418.1 | 38.3 | 9.2% | 458.7 | 119.8 | 26.1% |
| F5 | 406.7 | 14.0 | 3.5% | 408.0 | 13.7 | 3.4% |
| F6 | 403.2 | 7.7 | 1.9% | 404.1 | 9.2 | 2.3% |

## 4.3 Comparison with state-of-the-art algorithms

To comprehensively verify the efficiency of RPHSA in solving high-dimensional expensive problems, we compared it with three state-of-the-art algorithms, including ESAO [34], SA-COSO [35], and SHPSO [36], by testing them on all the six functions of 100 and 200 dimensions. Besides ESAO, SA-COSO and SHPSO also belong to hierarchical SAEAs, and have been briefly introduced in Section 2. To ensure the fairness of the comparison, the results of the three competitors were all obtained by running their source codes with recommended parameter settings. Tables 4 and 5 report the final optimization results generated by the total four algorithms on 100-D and 200-D functions, respectively, where the best results are highlighted in bold. Moreover, to further demonstrate the performance differences among the four algorithms, we present their convergence curves on 100-D and 200-D functions in Figs. 4 and 5, respectively. Note that the performance of SHPSO on 200-D problems was not investigated in the original paper [36], so we did not compare it with RPHSA on 200-D functions. According to the results in Tables 4 and 5 and Figs. 4 and 5, the following observations can be made:

**Table 4** Optimization results of the four algorithms on 100-D functions

| Function | Algorithm | Mean | Std. | Wilcoxon test |
|---|---|---|---|---|
| F1 | ESAO | 1.2781E+03 | 1.0345E+02 | + |
|  | SA-COSO | 9.6592E+02 | 2.6741E+02 | + |
|  | SHPSO | 1.1873E+02 | 3.3378E+01 | + |
|  | RPHSA | **3.4789E+01** | **1.1314E+01** |  |
| F2 | ESAO | 5.6681E+02 | 4.1628E+01 | + |
|  | SA-COSO | 2.5705E+03 | 8.7115E+02 | + |
|  | SHPSO | 2.0006E+02 | 3.9209E+01 | + |
|  | RPHSA | **1.2014E+02** | **2.1860E+01** |  |
| F3 | ESAO | 1.0469E+01 | **2.8768E-01** | + |
|  | SA-COSO | 1.5810E+01 | 6.5659E-01 | + |
|  | SHPSO | 5.4680E+00 | 7.3994E-01 | + |
|  | RPHSA | **3.0601E+00** | 2.9360E-01 |  |
| F4 | ESAO | 5.7359E+01 | 4.0934E+00 | + |
|  | SA-COSO | 6.0658E+01 | 1.5995E+01 | + |
|  | SHPSO | **1.1169E+00** | **4.9572E-02** | - |
|  | RPHSA | 1.7949E+00 | 2.5087E-01 |  |
| F5 | ESAO | **7.2524E+02** | **3.4544E+01** | - |
|  | SA-COSO | 1.2682E+03 | 1.2543E+02 | + |
|  | SHPSO | 8.1597E+02 | 5.0934E+01 | + |
|  | RPHSA | 7.8263E+02 | 6.7187E+01 |  |
| F6 | ESAO | 1.3703E+03 | 2.8016E+01 | + |
|  | SA-COSO | 1.3657E+03 | 2.2049E+01 | + |
|  | SHPSO | 1.4191E+03 | **1.5606E+01** | + |
|  | RPHSA | **1.3502E+03** | 2.0771E+01 |  |

1) RPHSA demonstrates the overall best performance. It obtains the best results on four out of total six 100-D functions, and is only defeated by SHPSO and ESAO on F4 and on F5, respectively. For 200-D functions, RPHSA shows an edge over all its competitors except being outperformed by SA-COSO on F5. Specifically, RPHSA wins significant superiority over other algorithms on F1-F3 of both 100-D and 200-D, demonstrating powerful capability of optimizing high-dimensional expensive problems. In conclusion, RPHSA could definitely be seen as the champion on this set of functions.

2) RPHSA has a good scalability. When the function dimension changes from 100 to 200, the performance of all the four algorithms degenerates to a certain extent. Despite this fact, RPHSA has the smallest performance degeneration and shows more obvious superiority over its competitors on most functions. The good scalability of RPHSA should be mainly attributed to the dimension reduction capability of RP.

3) RPHSA converges more stably and rapidly. It can be seen from Figs. 4 and 5 that no matter the function dimension is 100 or 200, RPHSA always keeps a more stable improvement tendency and continuously finds new better solutions during the whole evolution process. Its convergence rate is also faster than those of other algorithms on most functions, leading to better final optimization results.

**Table 5** Optimization results of the three algorithms on 200-D functions

| Function | Algorithm | Mean | Std. | Wilcoxon test |
|---|---|---|---|---|
| F1 | EASO | 1.7915E+04 | 1.1990E+03 | + |
|  | SA-COSO | 1.7042E+04 | 2.7424E+03 | + |
|  | RPHSA | **1.6149E+03** | **2.3990E+02** |  |
| F2 | EASO | 4.4657E+03 | 2.8440E+02 | + |
|  | SA-COSO | 1.6947E+04 | 4.5846E+03 | + |
|  | RPHSA | **6.2849E+02** | **8.3853E+01** |  |
| F3 | EASO | 1.4748E+01 | **2.1930E-01** | + |
|  | SA-COSO | 1.8073E+01 | 3.1409E-01 | + |
|  | RPHSA | **6.3853E+00** | 2.3864E-01 |  |
| F4 | EASO | 5.6048E+02 | 3.6043E+01 | + |
|  | SA-COSO | 5.4881E+02 | 9.3034E+01 | + |
|  | RPHSA | **7.4753E+01** | **1.3687E+01** |  |
| F5 | EASO | 5.3374E+03 | 1.7613E+02 | + |
|  | SA-COSO | **4.1770E+03** | 2.1812E+02 | - |
|  | RPHSA | 4.9156E+03 | **1.1404E+02** |  |
| F6 | EASO | 1.4495E+03 | **1.7449E+01** | + |
|  | SA-COSO | 1.3765E+03 | 1.8937E+01 | + |
|  | RPHSA | **1.2897E+03** | 4.9113E+01 |  |

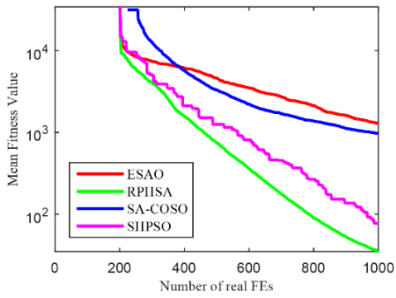

(a) F1

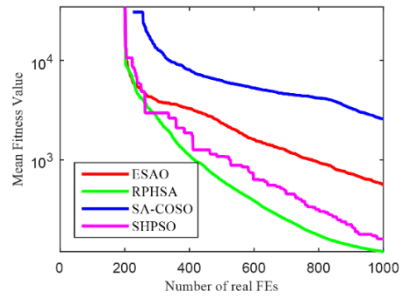

(b) F2

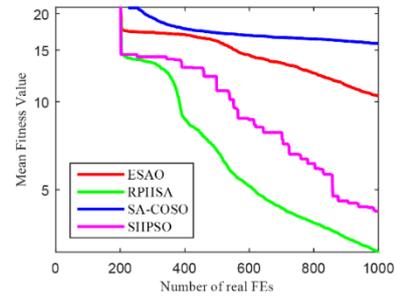

(c) F3

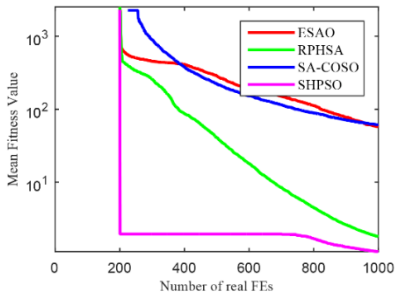

(d) F4

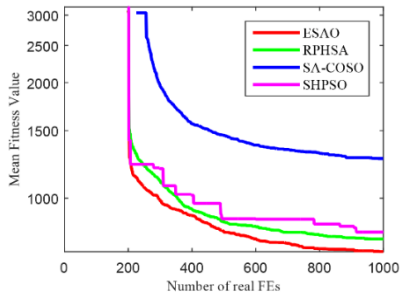

(e) F5

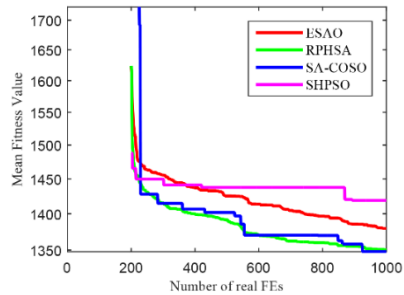

(f) F6

**Fig. 4** Convergence curves of the four algorithms on 100-D functions.

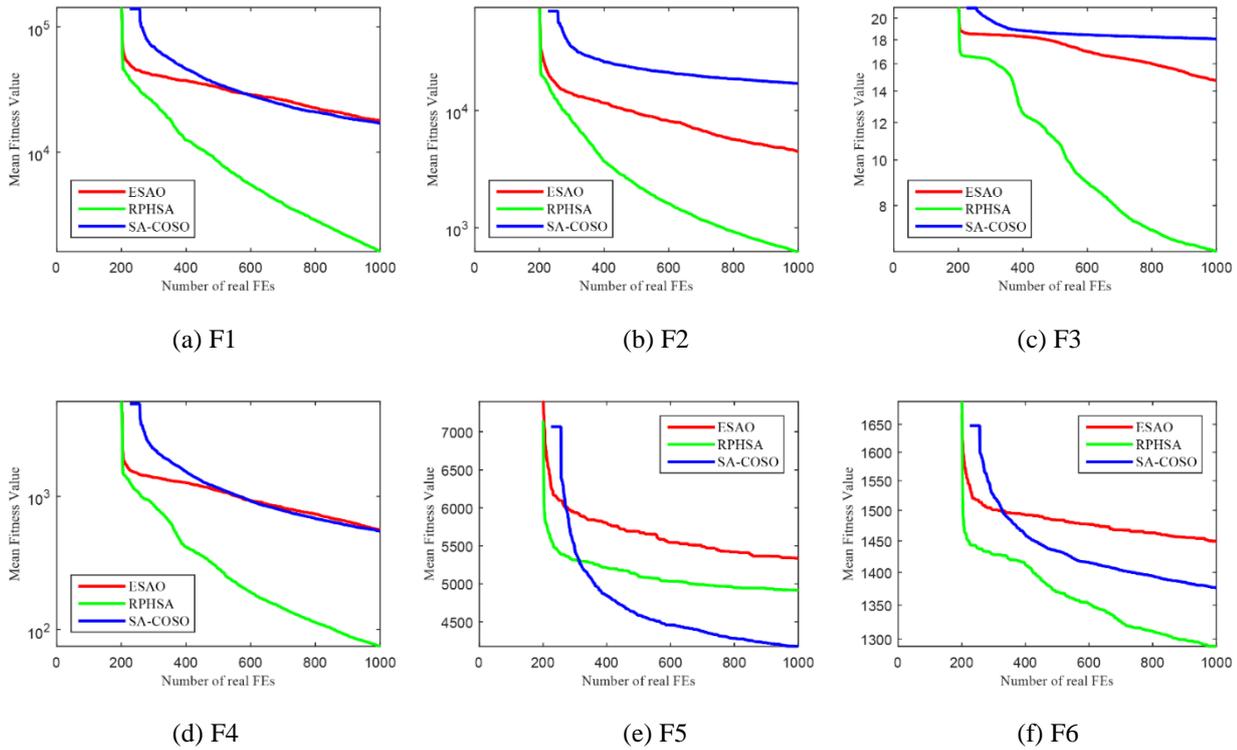

**Fig. 5** Convergence curves of the three algorithms on 200-D functions.

## 5. Conclusion

In this paper, an RP-enhanced hierarchical SAEA, i.e., RPHSA, is proposed to solve high-dimensional computationally expensive optimization problems. RPHSA inherits the fine framework of ESAO, but builds the local surrogate model therein, i.e., RBF, with the help of RP and develops a new local model named RP-RBF. Different from ESAO which directly trains its local model in the original high-dimensional space, RPHSA trains a group of RBFs in their respective subspaces generated by RP and constructs the final RP-RBF by averaging all low-dimensional RBFs. With the introduction of RP, not only the main characteristics of the original high-dimensional problem can be preserved to a large extent, but also the number of samples required for modeling can be significantly reduced.

Experimental results on six 100-D and 200-D functions indicate that RP-RBF has higher approximation accuracy and thus greatly enhances the optimization capability of RPHSA. Compared with three state-of-the-art SAEAs, RPHSA presents obvious superiority in solution quality, scalability, and convergence performance.

In future work, we will attempt to apply RPHSA to solve some real-world high-dimensional expensive problems. It is also interesting to scale up other kinds of surrogate models such as GP with RP under the framework of RPHSA.